\documentclass[letterpaper]{article} 
\usepackage[submission]{aaai25}  
\usepackage{times}  
\usepackage{helvet}  
\usepackage{courier}  
\usepackage[hyphens]{url}  
\usepackage{graphicx} 
\usepackage{natbib}  
\usepackage{caption} 
\usepackage{amsmath}
\usepackage{algorithm}
\usepackage{algorithmic}
\usepackage{lipsum}
\usepackage{booktabs}
\usepackage{colortbl}
\usepackage{threeparttable}
\definecolor{lightred}{rgb}{1.0, 0.8, 0.8}
\definecolor{lightgreen}{rgb}{0.8, 1.0, 0.8}
\definecolor{lightgray}{rgb}{0.8, 0.8, 0.8}
\usepackage{amsmath}
\usepackage{pifont}

\usepackage{newfloat}
\usepackage{listings}
\DeclareCaptionStyle{ruled}{labelfont=normalfont,labelsep=colon,strut=off} 
\floatstyle{ruled}
\newfloat{listing}{tb}{lst}{}
\floatname{listing}{Listing}
\title{De-biased Multimodal Electrocardiogram Analysis}
\author {
    Haitao Li\textsuperscript{\rm 1},
    Ziyu Li\textsuperscript{\rm 1},
    Yiheng Mao\textsuperscript{\rm 1},
    Ziyi Liu\textsuperscript{\rm 2},
    Zhoujian Sun\textsuperscript{\rm 3},
    Zhengxing Huang\textsuperscript{\rm 1},
}
\affiliations {
    \textsuperscript{\rm 1}Zhejiang University\\
    \textsuperscript{\rm 2}Transtek Medical Electronics Co., Ltd\\
    \textsuperscript{\rm 3}Zhejiang Lab\\
    lihaitao@zju.edu.cn, zhengxinghuang@zju.edu.cn
}

\usepackage{bibentry}

\begin{document}

\maketitle

\begin{abstract}
Multimodal large language models (MLLMs) are increasingly being applied in the medical field, particularly in medical imaging. However, developing MLLMs for ECG signals, which are crucial in clinical settings, has been a significant challenge beyond medical imaging. 
Previous studies have attempted to address this by converting ECGs into several text tags using an external classifier in a training-free manner. However, this approach significantly compresses the information in ECGs and underutilizes the reasoning capabilities of LLMs. In this work, we directly feed the embeddings of ECGs into the LLM through a projection layer, retaining more information about ECGs and better leveraging the reasoning abilities of LLMs. 
Our method can also effectively handle a common situation in clinical practice where it is necessary to compare two ECGs taken at different times.
Recent studies found that MLLMs may rely solely on text input to provide answers, ignoring inputs from other modalities. We analyzed this phenomenon from a causal perspective in the context of ECG MLLMs and discovered that the confounder, severity of illness, introduces a spurious correlation between the question and answer, leading the model to rely on this spurious correlation and ignore the ECG input. Such models do not comprehend the ECG input and perform poorly in adversarial tests where different expressions of the same question are used in the training and testing sets. We designed a de-biased pre-training method to eliminate the confounder's effect according to the theory of backdoor adjustment. 
Our model performed well on the ECG-QA task under adversarial testing and demonstrated zero-shot capabilities. An interesting random ECG test further validated that our model effectively understands and utilizes the input ECG signal.
\end{abstract}

%

\section{Introduction}
Electrocardiograms (ECGs) capture the heart's electrical activity, aiding physicians in diagnosing heart conditions such as arrhythmias and myocardial infarctions. With the advancement of deep learning, an increasing number of automated ECG analysis and diagnostic methods \cite{kiranyaz2015real, acharya2018automated, acharya2017deep, yildirim2018novel, tan2018application, guo2019inter} have emerged. Recently, multimodal large language models \cite{liu2023visual, alayrac2022flamingo} have gained significant attention, showing initial applications in medical imaging \cite{tu2023towards, li2023llava, moor2023med}. This raises the question: Can a multimodal large language model be developed for electrocardiograms? Unlike the image domain, which benefits from several well-established pre-trained encoders like CLIP \cite{radford2021learning} and SAM \cite{kirillov2023segment}, the ECG domain lacks such mature infrastructure, posing unique challenges.

\begin{figure}[t]
    \centering
    \includegraphics[width=1.0\linewidth]{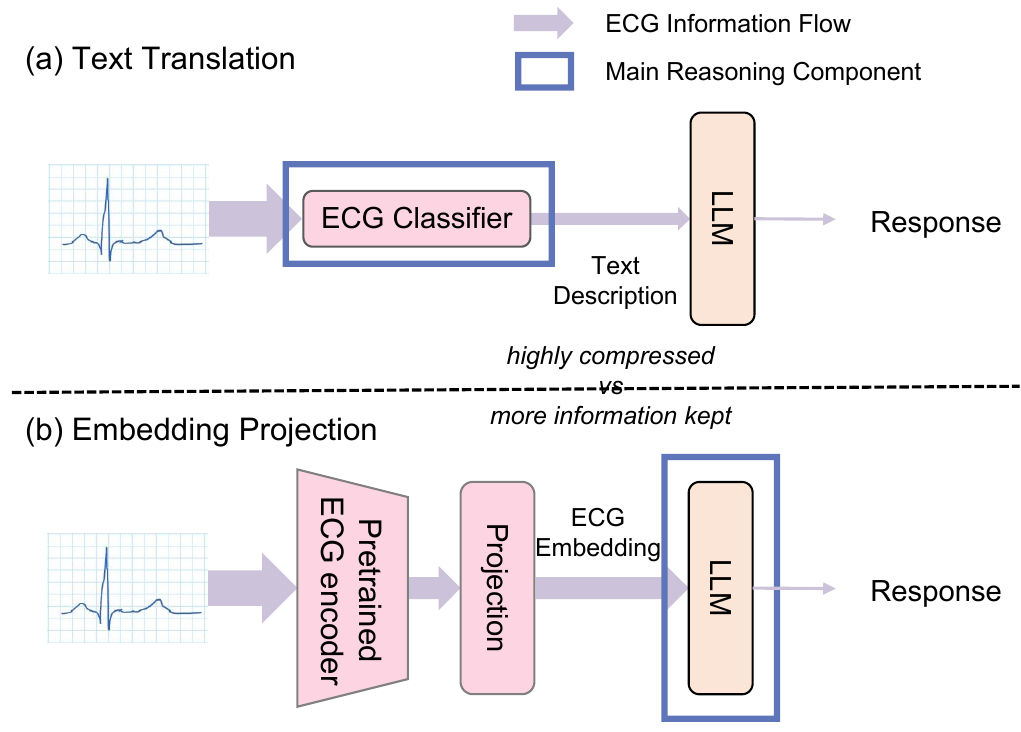}
    \caption{\textbf{Comparison of two methods for constructing ECG multimodal LLMs.} The ECG information fed into the LLM is highly compressed (text description) in the first approach compared to the second one (ECG embedding) and the first approach mainly reasons in external ECG classifiers while the second approach mainly reasons in the LLM. }
    \label{fig:intro_comparion}
\end{figure}

Due to the absence of large-scale pre-trained ECG encoders, some studies \cite{liu2023biosignal, oh2023ecg} have used traditional ECG classifiers to generate several tags for the ECG signal, thereby forming a text description. This text description is then employed to convey ECG information to the LLM. As illustrated in Figure \ref{fig:intro_comparion}, while this simple approach is training-free, representing ECG data as text description results in a significant loss of information. Moreover, the reasoning capabilities of LLMs are underutilized, as the primary ECG analysis relies on the external ECG classifier. Consequently, in this work, mimicking the approach used in visual language models(VLMs) \cite{liu2023visual, alayrac2022flamingo}, we pre-trained an ECG encoder using about 800,000 ECG-report pairs by multimodal contrastive learning \cite{radford2021learning}. We then map the ECG embedding encoded by the pre-trained encoder to the semantic space of the language model through projection, aiming to maximize the transmission of ECG information to the LLM and fully exploit its reasoning capabilities.

In clinical practice, ECGs are frequently compared to assess changes over time. Although some methods \cite{wan2024electrocardiogram} perform well in analyzing single ECG, they are not compatible with dual ECG signal comparison. In this study, we used simple prompts to instruct the LLM on which segment corresponds to the first ECG and which to the second, effectively accommodating this common medical scenario.

\begin{figure}[h]
    \centering
    \includegraphics[width=1.0\linewidth]{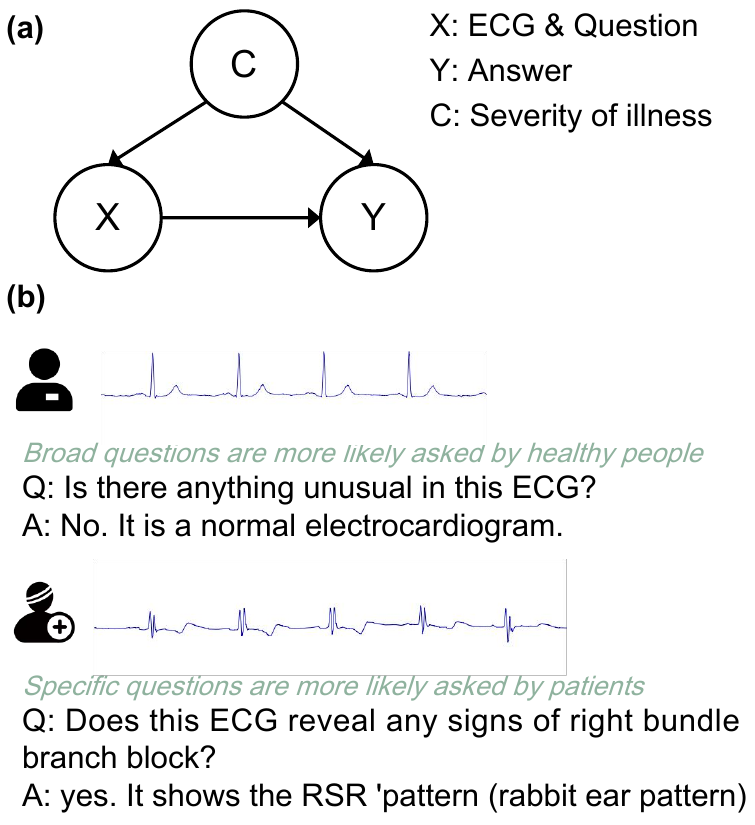}
    \caption{(a) \textbf{The Structural Causal Model of ECG Question-Answering}. The confounder `severity of illness' causes the spurious correlation X ← C → Y to mislead the model from the true objective X → Y. (b) Two examples illustrate this spurious correlation.}
    \label{fig:intro_bias}
\end{figure}

Additionally, recent studies have found that MLLMs may rely solely on text input to generate answers, ignoring inputs from other modalities. We investigated this issue from a causal perspective in the context of ECG MLLMs. We can formulate the causalities in the ECG question-answering process using a Structural Causal Model (SCM) \cite{pearl2016causal}. As shown in Figure \ref{fig:intro_bias}(a), the direct links denote causality between two nodes: cause → effect. X → Y represents the desired generation process of the target answer relying on ECG and question. However, we identified a confounder, the `severity of illness' (denoted as C), which influences both the ECG and the question (C → X) and the answer (C → Y), leading to a spurious association between X and Y. For C → X, as illustrated in Figure \ref{fig:intro_bias}(b), healthy individuals are more likely to ask broad questions due to their limited knowledge about the disease, whereas sick individuals tend to ask more specific questions because of their greater knowledge about the disease (we will prove this quantitatively in Table. \ref{tab: de-biased data}(Upper)). For C → Y, healthy individuals are more likely to receive answers without abnormalities. These two together lead to the spurious correlation between X and Y, that is, broad questions are more likely to have answers without abnormalities, while specific questions are more likely to have answers with serious diseases. As a result, The dataset (X, Y) is contaminated by the backdoor path X ← C → Y. So the model may rely on this spurious correlation to answer the question while ignoring the ECG input. Such models do not comprehend the ECG input and perform poorly in adversarial tests where different expressions of the same question are used in the training and testing sets. To eliminate the bias caused by confounder C and build a robust model, we designed a de-biased pre-training method based on the theory of backdoor adjustment \cite{pearl2016causal}.

Our contributions are as follows:
\begin{itemize}
    \item We developed an ECG MLLM by directly inputting ECG embeddings into the LLM through projection, maximizing the transfer of ECG information, and fully leveraging the LLM's reasoning capabilities. Additionally, our method is compatible with the common clinical practice of comparing two ECGs.
    \item We analyzed the incorrect biases in the ECG MLLM output generation process from a causality perspective and identified the confounder, `severity of illness', which led to spurious correlations between questions and corresponding answers. we designed a de-biasing pre-training method to address this issue.
    \item Experiments demonstrate the effectiveness of our method in the ECG-QA task under adversarial testing, and our model also exhibits outstanding zero-shot capabilities. An interesting Random ECG Test further verified that our model indeed understands and utilizes the input ECG signal.
\end{itemize}

\section{Related Work}
\textbf{Multimodal Large Language Models(MLLMs)}
With the continuous development and maturation of LLMs, attention has gradually shifted to MLLMs. Initially, vision language models such as BLIP-2 \cite{li2023blip}, LLaVa \cite{liu2023visual} and Flamingo \cite{alayrac2022flamingo} emerged. Subsequently, MLLMs encompassing other modalities like video \cite{maaz2023video, wang2023vaquita} and audio \cite{gong2023listen, rubenstein2023audiopalm} have also been developed. Researchers have been particularly interested in the applications of these MLLMs in the medical imaging domain, including models like Med-PaLM M \cite{tu2023towards}, LLaVa-Med \cite{li2023llava}, and Med-Flamingo \cite{moor2023med}. Recently, there have been efforts to develop MLLMs specifically for ECG analysis.

However, unlike the image domain, due to the lack of mature pre-trained ECG encoders, many studies \cite{liu2023biosignal, oh2023ecg} simply use a classifier to assign ECGs with a few tags, creating text descriptions and using these descriptions to convey ECG information to the LLM. For example, BioSignal Copilot \cite{liu2023biosignal} employs a suite of engines to convert biosignals into textual descriptions for easy processing by LLMs. Similarly, SE-WRN+LLM \cite{oh2023ecg} converts ECG signals into text descriptions using outputs from the ECG classifier, SE-WRN \cite{han2021towards} model, which provides probabilities for various common symptoms associated with the analyzed ECG. Although these methods are training-free, the disadvantages are also obvious as discussed in Figure \ref{fig:intro_comparion}. Therefore, we attempt to construct MLLMs for ECG analysis that directly inputs the ECG embeddings into the LLM to maximize information provision and fully utilize the LLM's reasoning capability.

\textbf{Robustness of MLLMs}
Although research on MLLMs is popular, some studies \cite{zhang2024visually} have found that despite their massive parameter sizes and high computational resource consumption, these MLLMs underperform on some simple classification tasks compared to traditional models. Additionally, other research \cite{rahmanzadehgervi2024vision} highlights that `vision language models are blind', VLMs fail to correctly identify whether two circles overlap, whether two lines intersect, and other relatively simple tasks. Moreover, it has been discovered that MLLMs lack robustness and cannot pass adversarial sample tests \cite{dong2023robust}. Specifically, in the context of medical imaging multimodal models, some studies indicate that these MLLMs perform worse than random under adversarial testing \cite{yan2024worse}. Some study \cite{liu2024paying} further explains this phenomenon from the scale disparity between the vision encoder and the language model, noting that since current VLMs are predominantly driven by the LLM, in extreme cases, VLMs may generate consistent descriptions with or without visual input, implying that certain outputs are influenced solely by the contextual text.

In this work, we aim to explore this phenomenon further within the context of ECG MLLM, specifically investigating whether in ECG Question-Answering, ECG MLLMs similarly rely on the question to produce answers while neglecting the input of the ECG. We seek to explain this issue from a causality perspective and find that due to the confounder of `the severity of illness', more specific questions are more likely to be posed by patients with illnesses, which in turn makes it more likely to have answers with an abnormal condition. This leads to a spurious correlation between the question and the answer, allowing the ECG MLLM to neglect the ECG signal and answer directly based on the question alone.

\section{Method}
In this section, we present a comprehensive description of our model architecture. Then, we analyze the spurious correlation induced by the confounder present in the dataset and explain in detail how we performed de-biased pre-training to mitigate the impact of the confounder.

\subsection{Architecture}
\begin{figure}[t]
    \centering
    \includegraphics[width=0.9\linewidth]{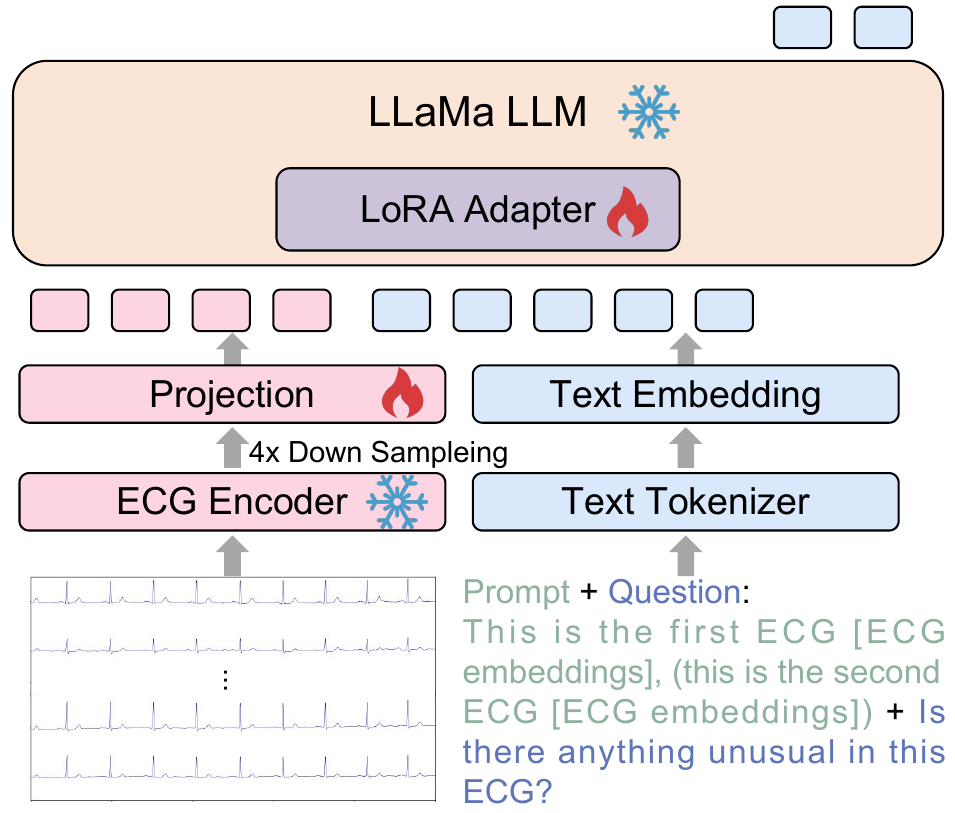}
    \caption{The overview of our model.}
    \label{fig:model}
\end{figure}

The architecture of our model is depicted in Figure \ref{fig:model}. It comprises an ECG encoder, a large language model, a modality alignment module, and LoRA adapters. We will explain each component in detail below.

\begin{table*}[ht]
    \centering
    \caption{(Upper). \textbf{Bias in the original dataset}. Broader questions are more likely to be raised by individuals with a normal ECG, and vice versa. (Lower). \textbf{The de-biased pre-training data we constructed}.}
    \label{tab: de-biased data}
    \begin{tabular}{lc}
    \toprule
    Question & Normal ECG Ratio\\ 
    \midrule
    Is there anything unusual in this ECG?	& 81.4\% \\
    Are there any abnormalities evident in the ECG?	& 79.1\% \\
    Does this ECG show any indications of abnormal cardiac activity? & 74.5\% \\
    \midrule
    Can the presence of \textbf{middle stage of myocardial infarction} be confirmed through this ECG? & 24.5\% \\
    Does this ECG reveal any signs of \textbf{incomplete right bundle branch block}? & 26.0\% \\
    Is \textbf{ischemic in anterior leads} indicated by this ECG? & 13.3\% \\
    Does this ECG reveal any signs of \textbf{non-diagnostic t abnormalities}? & 31.7\% \\
    \bottomrule
    \toprule
    \textbf{De-biased Pre-training Data: Paired yes and no data} \\
    \midrule
    ECG: \includegraphics[width=0.3\linewidth]{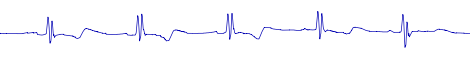} \\
    Question: Does this ECG reveal any signs of non-diagnostic t abnormalities? \\
    Answer: Yes \\
    \midrule
    ECG: \includegraphics[width=0.3\linewidth]{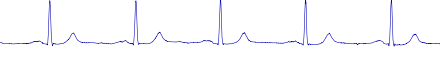} \\
    Question: Does this ECG reveal any signs of non-diagnostic t abnormalities? \\
    Answer: No \\
    \bottomrule
    \end{tabular}
\end{table*}

\noindent\textbf{ECG encoder}
We employed a ViT \cite{dosovitskiy2020image} modified for the characteristics of ECG signals, pre-trained on nearly 800,000 ECG-report pairs using multimodal contrastive learning \cite{radford2021learning}, as our ECG encoder. 

Considering an ECG signal denoted as $X \in {R}^{L \times T}$ with $L$ leads and length $T$, we first performed min-max normalization on the signal for each lead to eliminate measurement biases from different instruments and improve generalization capability. To adapt the Transformer \cite{vaswani2017attention} architecture, we need to patchify the ECG signal. However, considering that ECG diagnosis heavily relies on local features and that ECG datasets are far smaller than image datasets, we do not directly patchify the raw ECG signal. Instead, we patchify the feature maps obtained after applying two 1D convolution operations to the ECG signal. Specifically, given the raw ECG signal, we first apply two 1D convolutions with kernel sizes of 15 and 7, respectively, along with batch normalization and ReLU activation, to obtain the feature maps, which are then patchified. Generally, ECG patches can be obtained through three different methods: temporal, spatial, and spatio-temporal ways, as discussed in \cite{na2024guiding}. However, since we previously applied 1D convolution which treats the ECG signal as a multi-channel 1D signal, to maintain consistency, we used the temporal patch method. Specifically, the feature maps of $X$ after convolution are divided into non-overlapping patches defined as $\text{Patch} = \{\text{Patch}_1, ..., \text{Patch}_n\}$, where $n$ is determined by $T/p$, and $p$ represents the size of each patch. We also added a [CLS] token at the beginning to capture the global feature. These patches are then added to a learnable positional embedding, $\text{Pos} \in {R}^D$, to create an embedding sequence $E = \{E_1, ..., E_{n}\} \in {R}^{n \times D}$. This embedding sequence becomes the input for the Transformer encoder. 

The Transformer \cite{vaswani2017attention} encoder consists of alternating layers of multi-headed self-attention (MSA) and Feed Forward blocks. Layer normalization (LN) is applied before every block, and residual connections are used after every block. The Feed Forward contains two layers with a GELU non-linearity.

\begin{align}
z_0 &= \left[ \mathbf{CLS}; \mathbf{E}_1; \mathbf{E}_2; \cdots ; \mathbf{E}_n \right], \mathbf{E}_{\text{i}} \in {R}^{D}, \\
z'_\ell &= \text{MSA}(\text{LN}(z_{\ell-1})) + z_{\ell-1}, \quad \ell = 1, \ldots, L \\
z_\ell &= \text{Feed Froward}(\text{LN}(z'_\ell)) + z'_\ell, \quad \ell = 1, \ldots, L \\
\mathbf{y} &= \text{LN}(z^0_L)
\end{align}

To enhance ECG representation, we pre-trained our ECG encoder using multimodal contrastive learning, similar to the approach employed in vision-language models \cite{liu2023visual, wang2023cogvlm}. We utilized approximately 800,000 ECG-report pairs, following the method used in CLIP \cite{radford2021learning}. Specifically, we aligned paired ECGs with their corresponding reports while distinguishing them from unpaired reports to learn ECG representations.

\noindent\textbf{Large Language Model.}
In this paper, we use the LLaMA2-7B-chat as our LLM. LLaMA \cite{touvron2023llama} is pre-trained on a combination of natural language and programming language corpora in a self-supervised manner. The chat version, derived from LLaMA, is a fine-tuned model optimized for dialogue. Specifically, it is further trained on instruction-following prompts, enhancing its effectiveness for various reasoning and generation tasks.

\noindent\textbf{Modality Alighment.}
To enable the embedding of ECG can be fed into the LLM, similar to LLaVa \cite{liu2023visual}, we employed a simple linear projection to map the output of the pre-trained ECG encoder into the semantic space of the large model. To improve the efficiency of language model reasoning, we downsampled the output embedding of the ECG encoder by a factor of four before the projection. Specifically, we took the final layer output \( z_L \) of the ECG encoder, removed the [CLS] token, and performed the following down sampling on the remaining tokens:

\[
z[i] = concat(z[4i], z[4i+1], z[4i+2], z[4i+3]) \tag{5}
\]

\noindent\textbf{Low-rank Adapters.}
Due to the limited size of the ECG-QA dataset, to avoid overfitting and prevent LLM from catastrophic forgetting \cite{goodfellow2013empirical} which could lead to the loss of ability to answer general questions, we did not perform full parameter fine-tuning but instead used Low-Rank Adaptation \cite{hu2021lora}(LoRA). LoRA introduces a small set of learnable weights called LoRA adapters on top of the pre-trained LLM while keeping the language model's parameters unchanged. Each LoRA adapter is connected to a specific model layer and modifies its frozen parameters by adding a low-rank learnable matrix of the same size. In this study, we inject LoRA adapters (rank = 8 and $\alpha$ = 16) into the projection layers of the keys and queries in all self-attention layers \cite{vaswani2017attention} of the LLaMA model.

\subsection{Training}
In this part, we further analyze the spurious correlation in the dataset caused by the confounder `severity of illness' and provide a detailed explanation of how we trained our model.

\noindent\textbf{De-biased Pre-training.}
We have analyzed in Figure \ref{fig:intro_bias} that the bias in the traditional generative process \( P(Y | X) \) is introduced by the confounder: `severity of illness'. Now we need to perform the deconfounding using backdoor adjustment \cite{pearl2016causal} to obtain a debiased model \( P(Y|\text{do}(X)) \). Deconfounding seeks the true causal effect of one variable on another, and it is appealing to the ECG Question-Answering: given an ECG and Question \( X \), we hope \( Y \) responded by the model being faithful only to the content of the input \( X \) itself. The backdoor adjustment promotes the posterior probability \( P(Y | \text{do}(X)) \) from passive observation to active intervention as shown below:

\begin{equation}
P(Y | \text{do}(X)) = \sum_c P(Y | X, c)P(c) \tag{6}
\end{equation}

\noindent where \( c \) is the stratum for the confounder \( C \). This encourages the model to maximize \( P(Y | X, c) \) for every stratum \( c \), only subject to a prior \( P(c) \) listening to no one, and hence the model is deconfounded. 

We now apply Equation 6 to design the de-biased pre-training method for the confounder `severity of illness'. As discussed above in Figure \ref{fig:intro_bias}(a), the confounder `severity of illness' can affect both the ECG and Question (C → X) and the Answer (C → Y). Sick individuals tend to ask more specific questions due to their greater knowledge about the disease compared to healthy individuals (C → X). Additionally, sick individuals are more likely to receive answers indicating abnormalities (C → Y). These two factors together lead to a spurious correlation between X and Y, meaning that the more specific the question, the more likely the answer will indicate serious diseases. This spurious correlation can lead the model to learn incorrect dependencies, resulting in bias and causing it to ignore the ECG input, relying solely on the question to generate answers. From the Structural Causal Model (SCM) \cite{pearl2016causal}, `severity of illness' is the confounder in generating answers, causing the spurious correlation X ← C → Y to mislead the model from the true objective X → Y. We summarized the proportion of normal ECGs in both broad and specific questions in the dataset, as shown in Table \ref{tab: de-biased data}(Upper). It is shown that in broad questions, nearly 80\% are posed by individuals with normal ECGs. However, when the questions mention specific symptoms or diseases, the proportion of questioners with normal ECGs drops to less than 30\%.

Next, we introduce our de-biased pre-training method to eliminate the confounder. As shown in Equation 6, using the backdoor adjustment, we stratify the confounder, and train the model on each stratum. To mitigate the influence of this confounder `severity of illness', we constructed a de-biased dataset for pre-training, as illustrated in Table \ref{tab: de-biased data}(Lower). Specifically, we created a dataset containing only yes/no questions. For each question, we paired ECGs such that the answers were Yes and No, respectively. This approach ensures that for both specific and broad questions, there are paired positive and negative ECGs corresponding to Yes and No answers, thereby eliminating the influence of the severity of illness confounder. 

\noindent\textbf{Training Objective.}
Our model is trained on the next token prediction task, conditioned on past tokens and the reference ECG signal. Specifically, it maximizes \( P(x_t \mid x_{1:t-1}, \text{ECG}) \) through cross-entropy for all \( 1 < t \leq T \) given the text sequence \( x_{1:T} \) and the reference ECG.

In addition to the previously mentioned de-biased pre-training, we also need to conduct a second stage of training to enable the model to complete open-ended QA responses. In the second phase, we use more diverse questions, including multiple-choice(choose) and open-ended(query) questions, in addition to the judgment(verify) questions used in de-biased pre-training. For both stages, we froze the parameters of the ECG encoder and the LLM, training only the projection and LoRA adapters. More specific training settings for these two stages are shown in Table \ref{table: training_setting}. All experiments were conducted on one NVIDIA A800-80GB GPU.

\begin{table}[h]
  \centering
  \caption{The training setting.}
  \begin{tabular}{cccc}
  \toprule
  Stage & Task & LR & Epochs \\
  \midrule
  1 & de-biased verify & 1e-4 & 6\\
  2 & verify, choose, query & 2e-5 & 2\\
  \bottomrule
  \end{tabular}
\label{table: training_setting}
\end{table}

\begin{table*}[t]
    \centering
    \begin{threeparttable}
    \caption{Results of ECG-QA in different question types.}
    \label{tab: results-ecgqa}
    \begin{tabular}{cccccccc}
    \toprule
    Methods & S-Verify & S-Choose & S-Query & CC-Verify & CC-Query & CI-Verify & CI-Query \\
    \midrule
    M3AE & 74.6 & 57.1 & 41.0 & 75.5 & 20.1 & 75.3 & 4.2 \\
    MedViLL & 73.9 & 54.1 & 40.4 & 74.3 & 22.0 & 77.5 & 3.5 \\
    Fusion Transf. & 72.1 & 46.4 & 37.4 & 71.9 & 18.4 & 68.1 & 2.2 \\
    SE-WRN + gpt4 & 71.0 & 48.1 & 35.7 & 54.9 & 13.0 & 68.8 & 2.5 \\
    SE-WRN + gpt-3.5-turbo & 69.3 & 36.1 & 31.1 & 58.2 & 10.5 & 64.1 & 1.3 \\
    SE-WRN + text-davinci-003 & 75.0 & 37.8 & 36.0 & 56.3 & 15.4 & 71.5 & 1.4 \\
    \rowcolor{lightgray}Ours & \textbf{77.2} & \textbf{59.8} & \textbf{42.6} & \textbf{78.9} & 15.2 & \textbf{81.2} & 3.2 \\
    \bottomrule
    \end{tabular}
    \begin{tablenotes}
        \footnotesize
        \item S: Single, CC: Comparison-Consecutive, CI: Comparison-Irrelevant
    \end{tablenotes}
    \end{threeparttable}
\end{table*}

\begin{table*}[t]
    \centering
    \begin{threeparttable}
    \caption{Zero-Shot Evaluations.}
    \label{tab: zero-shot}
    \begin{tabular}{cccccccccc}
    \hline
    \toprule
    Methods & AF & LBBB & RBBB & LVH & RVH & SA & SB & SR & ST \\
    \midrule
    \textbf{CLIP}\\
    PTB-XL & 97.9 & 88.4 & 90.6 & 80.3 & 78.1 & 89.5 & 92.8 & 86.2 & 98.2 \\
    LUDB & 93.2 & 57.3 & 77.5 & 64.4 & 63.4 & 92.9 & 92.6 & 74.3 & 38.1 \\
    Shaoxing & 91.8 & 86.7 & 97.9 & 88.3 & 89.9 & \textbackslash{} & 99.4 & 95.2 & 99.3 \\
    CPSC & 97.8 & 97.7 & 94.2 & \textbackslash{} & \textbackslash{} & \textbackslash{} & \textbackslash{} & \textbackslash{} & \textbackslash{} \\
    \midrule
    \textbf{Ours}\\
    \rowcolor{lightred}\cellcolor{white}PTB-XL & 98.9 & 91.3 & 92.5 & 83.0 & 81.2 & 91.7 & \cellcolor{lightgreen}91.9 & 88.2 & 98.5 \\
    \rowcolor{lightred}\cellcolor{white}LUDB & 95.5 & 61.0 & 81.2 & \cellcolor{lightgreen}62.2 & 64.4 & \cellcolor{lightgreen}90.2 & 93.1 & 80.1 & 45.9 \\
    \rowcolor{lightred}\cellcolor{white}Shaoxing & 92.0 & 88.3 & 98.5 & 89.2 & 90.2 & \cellcolor{white}\textbackslash{} & 99.5 & 97.8 & 99.6 \\
    CPSC & \cellcolor{lightgreen}97.1 & \cellcolor{lightred}98.5 & \cellcolor{lightred}96.0 & \textbackslash{} & \textbackslash{} & \textbackslash{} & \textbackslash{} & \textbackslash{} & \textbackslash{} \\
    \bottomrule
    \end{tabular}
    \begin{tablenotes}
        \item \scriptsize AF: Atrial Fibrillation, LBBB: Left Bundle Branch Block, RBBB: Right Bundle Branch Block, LVH: Left Ventricular Hypertrophy, RVH: Right Ventricular Hypertrophy, SA: Sinus Arrhythmia, SB: Sinus Bradycardia, SR: Sinus Rhythm, ST: Sinus Tachycardia
    \end{tablenotes}
    \end{threeparttable}    
\end{table*}

\noindent\textbf{Two ECGs Comparison}
Comparing ECGs from different times is a common practice in clinical settings, and our method is well-suited to support this scenario. As depicted in Figure \ref{fig:model}, ECG embeddings are concatenated with text embeddings as input to the LLM. This approach allows for flexibility in the number of input ECGs. For a single ECG input, the prompt used is: `This is an electrocardiogram. [ECG embeddings].' For two ECG inputs, the prompt is: `Here are two electrocardiograms. The first is [ECG1 embeddings], and the second is [ECG2 embeddings].' Finally, the question is appended: `Answer the following questions based on the above content: [question]' to create the final input for the model. 

\begin{figure*}[t]
    \centering
    \includegraphics[width=0.96\linewidth]{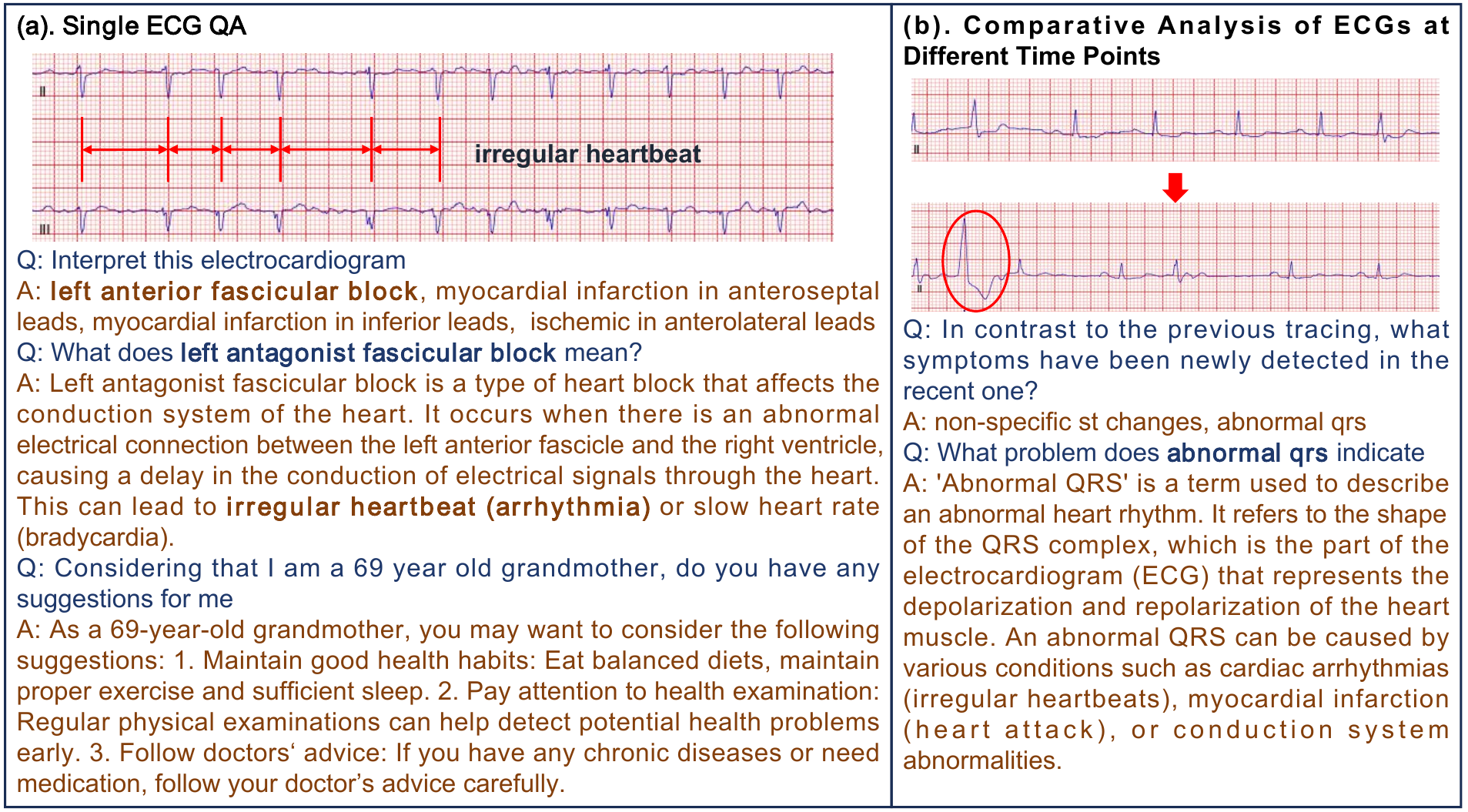}
    \caption{Case study}
    \label{fig: case_study}
\end{figure*}

\section{Experiments}
\subsection{Datasets}
We used two datasets to train our model. As discussed above, we first used MIMIC-ECG \cite{gowmimic} to pre-train our ECG encoder through multimodal contrastive learning like CLIP\cite{radford2021learning}. This dataset contains nearly 800,000 ECG-Report pairs with every ECG recording sampled at 500 Hz for a duration of 10 seconds. We removed samples with empty reports or reports containing fewer than three words to enhance data quality and resampled the ECG data to 100 Hz to improve training and inference efficiency.

ECG-QA \cite{oh2023ecg} dataset was used to train our ECG multimodal LLM. It contains 414,348 ECG question-answer pairs that include a variety of question types: Verify, Choose, and Query. In addition to single ECG question-answering, the dataset also includes dual ECG comparisons from the same person (Comparison-Consecutive) and different individuals (Comparison-Irrelevant). Additionally, it is also an adversarial test set, where the questions in the test set are different expressions of those in the training set which can be used to effectively test whether the model truly understands and utilizes the ECG input or simply relies on the text input to answer questions.

Simultaneously, we utilized the ECG-QA dataset to construct the de-biased dataset as discussed above. Specifically, we first selected Verify questions from the training set of ECG-QA. Then, for each question, we selected pairs of ECGs that provided answers of `yes' and `no'. This approach allowed us to create a dataset unaffected by the confounder `severity of illness', achieving a de-biasing effect.

\subsection{ECG-QA Result}
We validated the effectiveness of our method on the ECG-QA dataset, with the results presented in Table \ref{tab: results-ecgqa}. The evaluation metric used was Exact Match Accuracy, where an answer is considered correct only if the model's response exactly matches the standard answer. We compared our method against both traditional and LLM-based approaches. Traditional methods, such as M\textsuperscript{3}AE \cite{chen2022multi}, MedViLL \cite{moon2022multi}, and Fusion Transformer, treat the QA problem as a multi-label classification task, providing all possible answers as labels to the model, which limits these methods to closed-ended questions. Additionally, we compared our method with LLM-based approaches. As previously discussed, unlike our method, SE-WRN+LLM \cite{oh2023ecg} employs an external ECG classifier to assign tags to the ECG, which are then used as text input for the LLM. This strategy results in a significant loss of ECG information and underutilizes the reasoning capabilities of LLMs. As discussed above in Datasets, ECG-QA can be further divided into 7 subsets according to question type. The results demonstrate that, despite traditional methods having prior knowledge of possible answers, our model consistently outperformed both traditional and LLM-based methods across most subsets of the ECG-QA dataset.

\subsection{Zero-Shot Ability}
Inspired by the remarkable zero-shot capabilities demonstrated by LLMs, we conducted zero-shot evaluations on four previously unseen datasets: PTB-XL \cite{wagner2020ptb}, LUDB \cite{kalyakulina2020ludb}, the arrhythmia database from Shaoxing People's Hospital \cite{zheng2022large}, and CPSC \cite{liu2018open}. These datasets were originally designed for multi-label classification, from which we selected nine common diseases and symptoms, such as atrial fibrillation and sinus arrhythmia, to assess our model's zero-shot capabilities.

Specifically, for each disease or symptom, we evaluated the model by querying our ECG MLLM with the prompt, `Does this ECG reveal any signs of [label]?' The results, presented in Table \ref{tab: zero-shot}, were evaluated using the AUC metric. We compared our model with the above ECG model trained on ECG-report pairs from the MIMIC-ECG \cite{gowmimic} dataset using multimodal contrastive learning like CLIP \cite{radford2021learning}. Our model demonstrated superior zero-shot performance across the majority of labels, underscoring its effectiveness in handling zero-shot scenarios and highlighting its potential for robust performance across diverse datasets and medical conditions.

\subsection{Ablation Study with Random ECG Test}
We conducted ablation experiments on our de-biased pre-training method and designed a simple yet effective Random ECG Test to verify whether our ECG MLLM indeed understands and utilizes the ECG signal. In the Random ECG Test, we replaced the original ECG in the question with a random ECG and then observed the change in accuracy. A significant decrease in accuracy would indicate that the model effectively leverages ECG information and that the random ECG misled the model, causing the accuracy to drop. Conversely, minimal changes in accuracy would suggest that the model did not effectively use the ECG data, likely relying solely on the question to generate answers. We applied the Random ECG Test to models with and without de-biased pre-training, and the results are presented in Table \ref{tab: ablation}. These results can be analyzed from three perspectives:

\begin{table}[h]
    \centering
    \small
    \begin{threeparttable}        
    \caption{Ablation study and Random ECG Test.}
    \label{tab: ablation}
    \begin{tabular}{cccccl}
    \toprule
    De-bias & RET & S-Verify & CC-Verify & CI-Verify & \textbf{Avg.} \\ 
    \midrule
    \ding{51} & \ding{55} & 77.2 & 78.9 & 81.2 & \textbf{79.1}\\ 
    \ding{51} & \ding{51} & 63.6 & 61.9 & 57.2 & \textbf{60.9}\textsubscript{$\downarrow18.2$}\\
    \midrule
    \ding{55} & \ding{55} & 69.2 & 64.7 & 66.8 & \textbf{66.9}\\
    \ding{55} & \ding{51} & 66.5 & 63.6 & 64.5 & \textbf{64.9}\textsubscript{$\downarrow2.0$}\\
    \bottomrule
    \end{tabular}
    \begin{tablenotes}
        \footnotesize
        \item RET: Random ECG Test
    \end{tablenotes}
    \end{threeparttable}
\end{table}

\begin{enumerate}
    \item The significant performance drop in our de-biased model under the Random ECG Test shows that the model effectively understands and relies on ECG information.
    \item In contrast, the model without de-biased pretraining shows little performance change after random ECG replacement, indicating it did not effectively utilize ECG data and likely relied on the spurious correlations mentioned earlier to answer based solely on the question.
    \item The de-biased model's accuracy under the Random ECG Test is even lower than the non-de-biased model(about Random Guess), suggesting that the random ECG misled the model, causing the accuracy to be lower than Random Guess, confirming the understanding of ECG.
\end{enumerate}

\subsection{Case Study}
We present two case studies to demonstrate the effectiveness of our approach in single ECG QA and dual ECG comparison QA, as shown in Figure \ref{fig: case_study}. Compared to traditional models, our method brings powerful interactive capabilities, specifically characterized by flexible input and output. In addition to the ECG signal, users can provide the model with other auxiliary information such as medical history, gender, age, etc., to assist in diagnosis. Moreover, besides the diagnostic results, the model can clarify and explain any uncertainties users may have regarding these outcomes.

\section{Conclusion}
In this work, we developed an ECG MLLM that directly processes ECG embeddings, fully providing ECG information to the LLM and leveraging its reasoning capabilities. Our approach is also compatible with common clinical scenarios that involve comparing two ECGs. Additionally, we analyzed the biases in the ECG MLLM output generation process from a causality perspective, attributing these biases to the confounder: the severity of illness. To address this issue, we designed a de-biasing pre-training method based on the theory of backdoor adjustment. Extensive experiments demonstrate that our model exhibits outstanding performance on the ECG-QA task under adversarial testing and also shows strong zero-shot capabilities. The Random ECG Test further validated that our multimodal model effectively understands and utilizes ECG information.

\bibliography{reference}

\clearpage

\end{document}